\documentclass[runningheads]{llncs}
\usepackage{subcaption}
\usepackage{float}

\usepackage{graphicx}
\usepackage {xcolor}
\usepackage{amsfonts}
\usepackage{amsmath}

\usepackage[T1]{fontenc}
\def\doi#1{\href{https://doi.org/\detokenize{#1}}{\url{https://doi.org/\detokenize{#1}}}}

\graphicspath{ {./images/} }
\usepackage{ulem}

\begin{document}
\title{ColNav: Real-Time Colon Navigation for Colonoscopy}

%
%
\author{Netanel Frank$^\ast$  \and
Erez Posner$^\ast$  \orcidID{0000-0003-2778-1612} \and
Emmanuelle Muhlethaler \orcidID{0000-0003-2766-6999}\and
Adi Zholkover \and Moshe Bouhnik \orcidID{0000-0003-3003-2604}}
%
\authorrunning{E. Posner et al.}
\def\thefootnote{*}\footnotetext{These authors contributed equally to this work}\def\thefootnote{\arabic{footnote}}

%
\institute{Intuitive Surgical, Inc. \\ 1020 Kifer Road, Sunnyvale, CA \\ \email{\{netanel.frank, erez.posner, emmanuelle.muhlethaler, adi.zholkover ,moshe.bouhnik\}@intusurg.com}}
\maketitle              
\begin{abstract}
Colorectal cancer screening through colonoscopy continues to be the dominant global standard, as it allows identifying pre-cancerous or adenomatous lesions and provides the ability to remove them during the procedure itself. Nevertheless, failure by the endoscopist to identify such lesions increases the likelihood of lesion progression to subsequent colorectal cancer. Ultimately, colonoscopy remains operator-dependent, and the wide range of quality in colonoscopy examinations among endoscopists is influenced by variations in their technique, training, and diligence. This paper presents a novel real-time navigation guidance system for Optical Colonoscopy (OC).
Our proposed system employs a real-time approach that displays both an unfolded representation of the colon and a local indicator directing to un-inspected areas. These visualizations are presented to the physician during the procedure, providing actionable and comprehensible guidance to un-surveyed areas in real-time, while seamlessly integrating into the physician's workflow. Through coverage experimental evaluation, we demonstrated that our system resulted in a higher polyp recall (PR) and high inter-rater reliability with physicians for coverage prediction. These results suggest that our real-time navigation guidance system has the potential to improve the quality and effectiveness of Optical Colonoscopy and ultimately benefit patient outcomes.


\end{abstract}

\section{Introduction}
Colorectal cancer (CRC) is a significant public health issue, with over 1.9 million new cases diagnosed globally in 2020~\cite{ccfs}. It is one of the most preventable types of cancer~\cite{ijcm113278}, and early detection is crucial for preventing its progression~\cite{zjrms93934,Holakouie-Naini}. The most commonly used screening method is optical colonoscopy (OC)~\cite{Liang2010}, which visually inspects the mucosal surface of the colon for abnormalities such as colorectal lesions. However, the process of detecting CRC in its early stages can be difficult, since performing a comprehensive examination of the colon using OC alone can be challenging, resulting in certain regions of the colon not being fully examined and potentially reducing the rate of polyp detection.

To address this problem, researchers have conducted extensive studies to propose assistive technologies that set out to provide clinicians with a better understanding of the procedure quality. Most existing methods focus on estimating measures of quality, such as the withdrawal time, or on reconstructing a 3D model of the colon from a video sequence of the procedure.
Despite the advancements in technology that allow for the prediction of 3D structures from images, there is still a significant gap in providing useful and actionable information to clinicians during the procedure in real-time. Current methods for detecting un-surveyed regions, which usually show a 3D visualization of the colon, are not designed to be easily understood, or interacted with, during the procedure. They may not align with the camera view; making it difficult for physicians to understand where they need to move the endoscope to survey missing regions. Other measures of quality, such as coverage per frame, or withdrawal time, do not provide clear, usable information to assist during the procedure in capturing un-surveyed regions.
In this paper, we present ColNav, a novel real-time solution that (i) utilizes an unfolded representation of the colon to localize the endoscope within the colon, (ii) introduces a local indicator that directs the physician to un-surveyed areas and (iii) is robust to real-life issues such as tracking loss. Our approach estimates the centerline and unfolds the scanned colon from a 3D structure to a 2D image in a way that not only calculates the coverage, but also provides augmented guidance to un-surveyed areas without disrupting the physician's workflow. To the best of our knowledge, this is the first coverage based, real-time guidance system for colonoscopies.
\begin{figure}[tp]
\centering
\includegraphics[width=0.8\textwidth, height=0.42\textwidth]{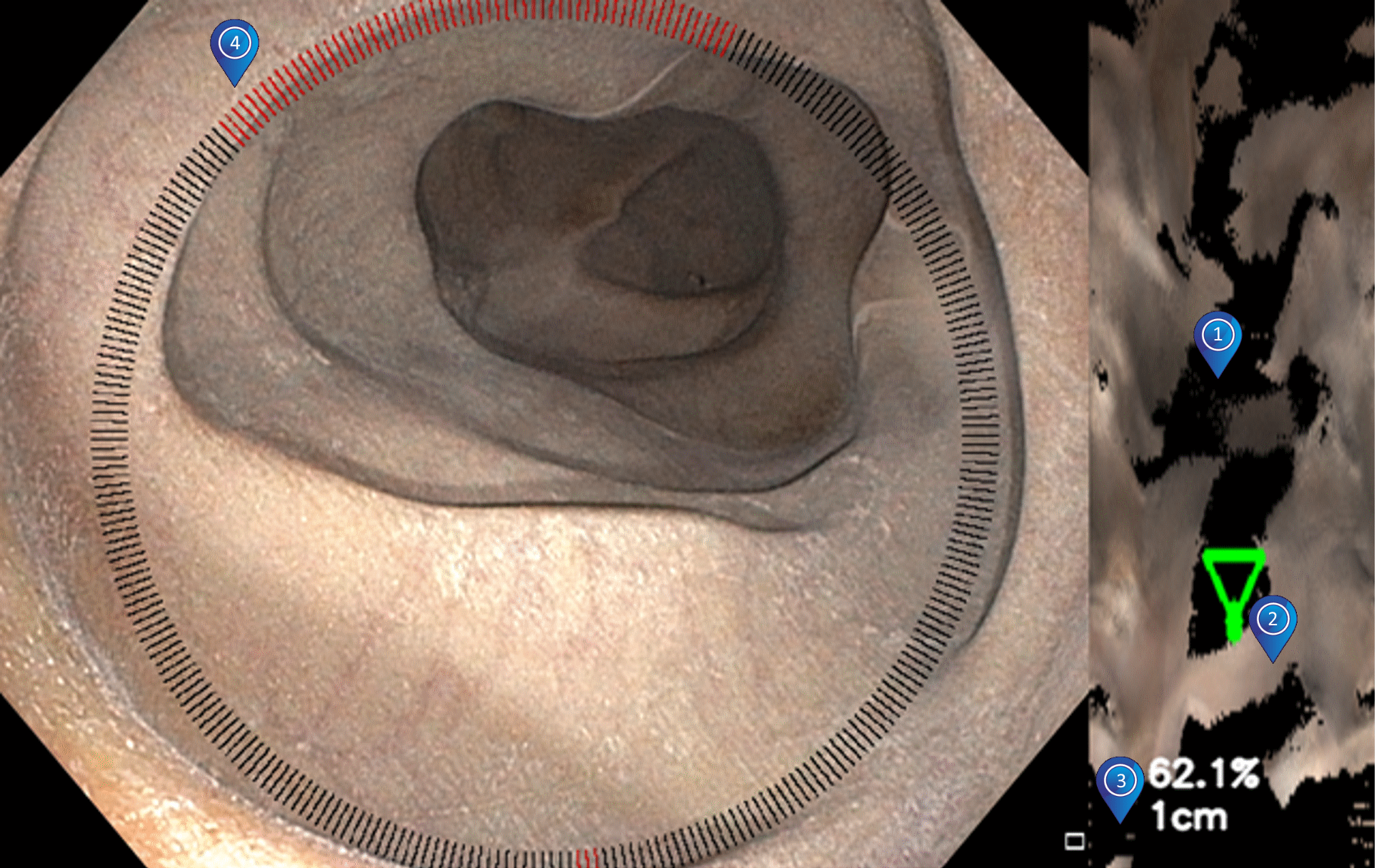}
\caption{Our novel, real-time colonoscopy navigation. Flattened image of the colon (right): (1) un-surveyed areas as black pixels, (2) camera location,  (3) coverage percentage and length covered. Endoscope view (left): (4) the local compass indicator directing the physician to look up (ticks highlighted in red).} \label{ui}
\end{figure}

\section{Related Work} 
\label{related_work}
In recent years, there has been an abundance of papers exploring various aspects of quality measures for colonoscopy, with the goal of assisting clinicians and improving the overall quality of care.

\textbf{SLAM for colonoscopy} \label{related_work_mono_slam}
 approaches usually rely on estimating a 3D reconstruction of the colon and post-processing it in order to estimate the un-surveyed regions (holes). Posner et al.~\cite{c3fusion} utilized deep features to better track the camera position, presented tracking loss recovery and loop closure capabilities to create a consistent 3D model. Ma et al.~\cite{RNNSLAM}  reconstructed fragments of the colon using Direct Sparse Odometry (DSO)~\cite{DSO} and a Recurrent Neural Network (RNN) for depth estimation. 
However their output is not easily understood nor meant to be interacted with during the procedure, making them less likely to be adopted by physicians or impact the clinical outcome.

\textbf{Direct coverage estimation} \label{related_work_coverage_direct}
methods~\cite{google_paper,9607561}, aim to predict the coverage on a segment-by-segment basis by estimating what fraction of the colon has been viewed in any given segment. Freedman et al.~\cite{google_paper} used a CNN to perform depth estimation for each frame followed by coverage estimation. 
As it was trained in a supervised manner using synthetic ground-truth coverage, it cannot be easily generalized to real data.
Blau et al.~\cite{9607561} proposed 
an unsupervised learning technique for detecting deficient colon coverage segments modeled as curved cylinders. However, their method does not run in real-time.

\textbf{Indirect Quality Objective measurements} \label{related_work_other}
Objective measurements of quality in colonoscopy are important for minimizing subjective biases and variations among endoscopists. Yao et al.~\cite{8856625} proposed an auto-detection of non-Informative frames and Zhou et al.~\cite{ZHOU2020428} predicted the bowel preparation scores every 30 seconds during the withdrawal phase of the procedure. However, these techniques lack the ability to provide real-time information to the physician during the procedure.

\textbf{Virtual colon unfolding}
\label{related_work_colon_unfolding}
is a well known visualisation technique for virtual colonoscopy (VC), where the colon is inspected by analysing the output of a CT scan. In such cases, a 3D mesh of the entire colon can be extracted from the CT image volume and mapped onto a 2D grid, providing the physician with a fast and convenient way to inspect the colon mucosa and find polyps. A number of solutions have been proposed to perform this mapping \cite{Wang98,Haker2000NondistortingFM,Vilanova2001NonlinearVC,Sudarsky2008}. In many cases, the solution uses, as an intermediate step, the computation of the \textit{centerline}, a single continuous line spanning the colon. Although most of these methods tend to be computationally expensive, Sudarsky et al.~\cite{Sudarsky2008} proposed a fast unfolding method based on the straightening of the colon mesh using the centerline. Using colon unfolding to visualize missed areas in optical colonoscopy (OC) has been proposed by Ma et al.~\cite{RNNSLAM}. However, it was done offline, for validation purposes, on a few disconnected colon segments, using a single straight line as the centerline.

\section{Method Overview}
\label{method_overview}
Our pipeline, ColNav, provides actionable and comprehensible guidance to un-surveyed areas in real-time and is seamlessly integrated into the physician’s workflow. While scanning the colon, the physician is presented with 2 screens as can be seen in Fig.~\ref{ui}. The colon's unfolded image, presented on the right, shows the three-dimensional (3D) colon flattened into a two-dimensional (2D) image. 
Black pixels in the image indicate unseen areas, which were missed during the scan. The location of the camera in the unfolded colon is visualized as the green camera frustum marker. When the physician withdraws the endoscope the green marker moves down and vice-versa. This enables the physician to know whether to move the endoscope forwards or backwards to reach the missed regions (holes). Coverage percentage and the overall length, computed on the scanned portion of the colon, are displayed as well. On the left, is the main endoscope view with a local compass indicator directing to un-inspected areas. Once the endoscope is near an un-inspected area (hole), the compass indicator directs the physician towards the area that needs to be examined.

The ColNav algorithm consists of three major parts: ($i$) centerline estimation, ($ii$) multi-segment 3D to 2D unfolding, and ($iii$) local indicator (navigation compass). To estimate the depth map and pose of each new frame, we employ C$^3$Fusion~\cite{c3fusion} as our SLAM module. ColNav's first component is a robust method for centerline estimation. In the second component, the overall 2D scene representation is obtained by merging the depth, pose, and RGB of all frames into a single flattened representation of the colon.
In real-life scenarios, C$^3$Fusion may lose track, resulting in the creation of a new segment when the last frame cannot be connected to any previous frame. Alternatively, loop closure may occur, where two disjoint segments are merged, and their poses are subsequently updated. Our system accommodates these scenarios by (a) adjusting the previous centerline approximation based on updated poses and (b) de-integrating frames that have changed location in the flattened image and re-integrating them with their new pose.
In cases of tracking loss, the flattened image shows separate segments with red lines that can be merged if tracking recovery occurs.

\subsection{Centerline \& Colon Unfolding}
\label{centerline}

\begin{figure}[tbp]
    \centering
    \begin{tabular}[t]{cc}

            \begin{subfigure}{.7\textwidth}
            \begin{tabular}{c}
                \smallskip

                    \begin{subfigure}[t]{1\textwidth}
                        \centering
                        \includegraphics[width=0.9\textwidth, height=.4\textwidth]{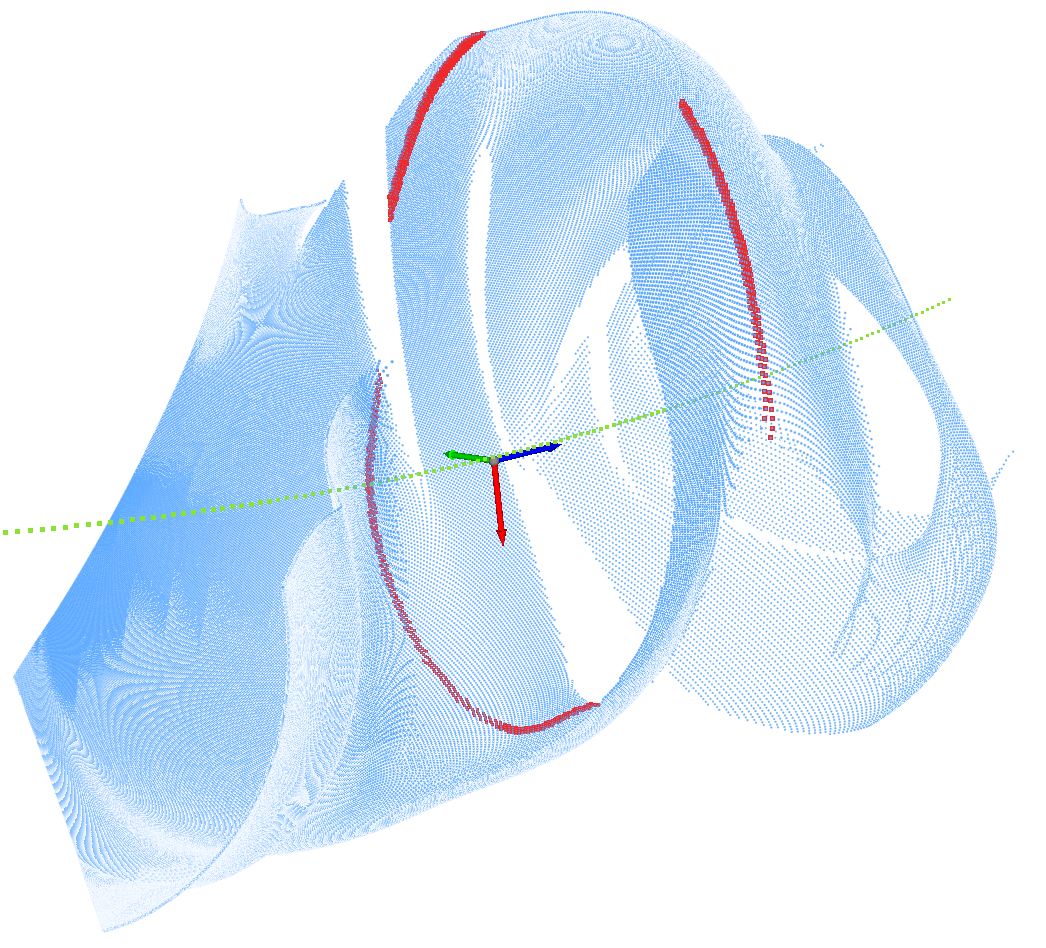}
                    \end{subfigure}
            \end{tabular}
                    \end{subfigure}

        \begin{subfigure}{.25\textwidth}
            \begin{tabular}{c}
                \smallskip
 
                    \begin{subfigure}[t]{1\textwidth}
                        \centering
                        \includegraphics[width=0.45\textwidth,height=1.1\textwidth]{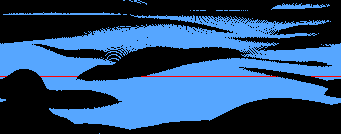}
                    \end{subfigure}
            \end{tabular}
        \end{subfigure}

    \end{tabular}

    \caption{On the left: 3D point cloud of a single frame (blue), centerline (green), vertices associated with a specific cross-section on the centerline (red) and the camera pose indicated by the 3 axis vectors. On the right: the flattened image with corresponding cross-section (red). Note that the holes in the cross-section match the black pixels in the corresponding row.} \label{centerline_fig}
\end{figure}

In our proposed solution, the 3D representation of each frame is obtained by back-projecting the depth map into a point cloud. It is then mapped onto a 2D  unfolded representation, using an algorithm analogous to that described in~\cite{Sudarsky2008}. In particular, the centerline is used for straightening the reconstructed colon and dividing it into cross-sections perpendicular to the centerline. Each cross-sectional slice corresponds to a row within the two-dimensional flattened image, see Fig.~\ref{centerline_fig}. The colon centerline, sometimes also referred to as the medial axis, is usually defined as a single connected line, spanning the colon and situated at its center, away from the colon walls~\cite{Wan2002}. In our case, the centrality requirement is partly relaxed, as we observed that a shift of the centerline away from the center of the colon has little effect on the unfolding. In our case, unlike prior works, the centerline and the flattened image are updated in real time. This creates new requirements for the centerline: (1) Fast computation. (2) Consistency over time relatively to the camera trajectory. To support these requirements, the centerline is estimated from the camera trajectory poses.

The centerline algorithm contains the following steps: (1) Filtering outlier poses from the trajectory. (2) Constructing or updating a graph $G$ of the trajectory, with camera positions as nodes and edges connecting nodes within a threshold distance. (3) Calculating or updating the shortest path length $l$ between each node $n\in G$. (4) Binning of the trajectory points according to $l$. (5) Fitting a B-spline~\cite{Dierckx1982} to an aggregate of the trajectory points in each bin. Each time the trajectory poses are updated, steps (1) - (5) are computed and a new centerline is re-calculated.

\textbf{Camera Position Indicator:}
The camera position for each frame is given by the SLAM module and is noted by ${T_i}=\{(R_i,t_i)|R_i \in SO(3),t_i \in \mathbb{R}^3\}_{i=1}^N$ with $N$ the number of frames in the sequence. To represent the endoscope current location $s_e$ on the centerline of size K, The endoscope position $t_i$ is projected on the centerline $C=\{c_k \in \mathbb{R}^3\}_{k=1}^K$ by querying the centerline KDTree.
\begin{equation}
    \begin{split}
    s_e = \arg \min_{c_k \in C}||t_i - c_k||^2 , t_i \in \mathbb{R}^3
    \end{split}
\end{equation}

\subsection{Navigation Compass}
The navigation compass serves as a local indicator that visually guides the physician to areas that have been missed. Specifically, the compass ticks are highlighted in red to indicate which specific sections of the colon require further inspection. Based on the camera position along the centerline $s_e$, the coverage information is extracted from the unfolded image $F$, where each column represents $b_{\theta}$ - the rotation angle bin around the centerline axis at the endoscope location $s_e$, with $\theta\in\{0,...,2\pi\}$. 

When $F(s_e,b_{\theta})$ - the corresponding pixel in the extracted row, is black (meaning, it wasn't covered), the navigation compass tick will be highlighted in red, otherwise it will remain dark.
To make the navigation compass invariant to camera roll, the camera orientation is projected on the centerline and the relative angle offset is computed to compensate for misalignment between the centerline and the camera pose.~Fig.~\ref{centerline_fig} depicts the extracted row, selected from the flattened image according to the camera location.


\subsection{Unfolding real-time dynamic update}
To achieve real-time and consistent unfolding of the colon, it is crucial to update the flattened image $F$ whenever new information becomes available. This need arises as the SLAM pipeline continually refines frames poses, updates frames segment assignment, and copes with real-life issues. To accomplish this, we closely monitor the continuous change in frames' poses and their assignment to segments, updating the flattened image through the integration and de-integration of frames. By adopting this strategy, we can rectify errors resulting from registration drift or tracking loss.

\textbf{Managing Unfolding Updates}
When an input frame arrives, we seek to integrate it into the
flattened image as quickly as possible, to give the physician instantaneous feedback of the colon coverage. Since previous frames segment assignment or poses could be updated from~\cite{c3fusion}, 
we de-integrate and re-integrate all frames if their segment assignment changes. In addition, we sort all frames within each segment in descending order, based on the difference between their previous and updated poses. After sorting, we select and re-integrate the top 10 frames from the list.
This allows us to dynamically update the unfolded image to produce
a globally-consistent representation of the unfolded colon.

\textbf{Integration and De-integration:}
Integration of an RGBD frame $f_i$ is defined as a weighted average of previous mapped samples. For each pixel $p$ in the flattened image, let $F(p)$ denote its color, $W(p)$ the pixel weight,
$d_i(p)$ the frame's sample color to be integrated, and $w_i(p)$ the integration weight for a sample of $f_i$. Each pixel is then updated by:

\begin{equation}
    F^{'}(p) = \frac{F(p)W(p) \pm w_i(p)d_i(p)}{W(p) \pm w_i(p)}, W^{'}(p) = W(p) \pm w_i(p)
\end{equation}
Where the $+$ sign is used for integration and the $-$ for de-integrating a frame.
A frame in the flattened image can be updated by de-integrating
it from its original pose and/or segment and integrate it with a new pose into its updated segment.

\section{Experiments}
\begin{figure}[ht]
    \centering
    \begin{tabular}[t]{cc}

        \begin{subfigure}{.5\textwidth}
            \begin{tabular}{c}
                \smallskip
 
                    \begin{subfigure}[t]{1\textwidth}
                        \centering
                        \includegraphics[width=0.8\textwidth,height=.5\textwidth]{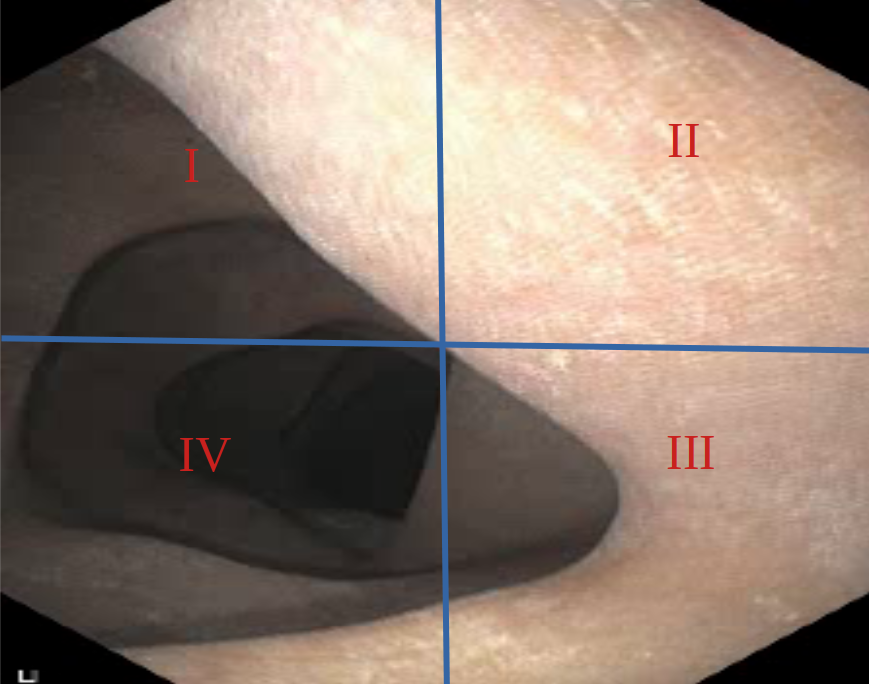}
                    \end{subfigure} \\
                    \begin{subfigure}[t]{1\textwidth}
                        \centering                        \includegraphics[width=0.8\textwidth,height=.1\textwidth]{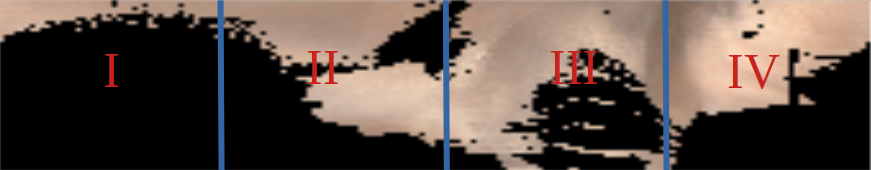}
                    \end{subfigure}
            \end{tabular}
        \end{subfigure}
        \begin{subfigure}{.5\textwidth}
            \begin{tabular}{c}
                \smallskip

                    \begin{subfigure}[t]{1\textwidth}
                        \centering
                        \includegraphics[width=0.8\textwidth,height=.625\textwidth]{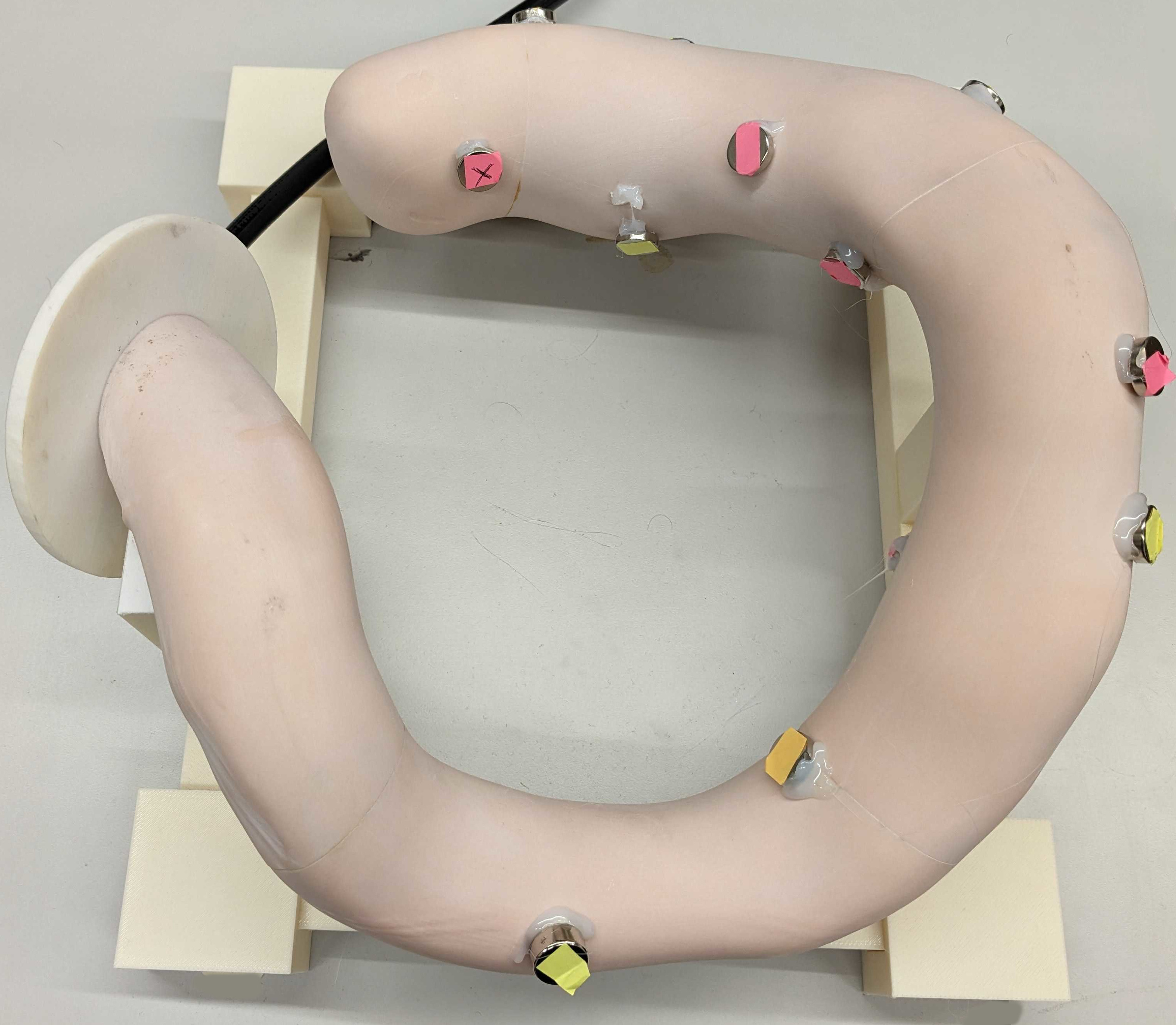}
                    \end{subfigure}
            \end{tabular}
        \end{subfigure}

    \end{tabular}

    \caption{Left: Representative frame from an annotated clip with the annotated quadrant numbers. Below, ColNav's flattened image of the same clip with the corresponding quadrant numbers. Note that areas that are occluded or aren't visible in the frame are mostly dark in the flattened image. Right: Our complete 3D model with the external magnets that hold in place the small magnetic balls.}
\label{fig:exps}
\end{figure}

This section presents the validation of our solution through multiple tests. The first test, named 'Colon unfolding verification', demonstrates that our 2D flattened visualization is a valid representation of the scanned colon. It also showcases our ability to detect and localize 'holes' in the colon using this visualization. To carry out this test, we used coverage annotations of short colonoscopy clips. Each clip was divided into four quadrants (See Fig.~\ref{fig:exps}), and two experienced physicians tagged each quadrant based on its coverage level ('mostly not covered', 'partially covered', 'mostly covered'). We then used ColNav to estimate the coverage of each quadrant and compared it to the physicians' annotations. The second test focuses on the clinical impacts of using our tool during procedures. We estimate coverage and Polyp Recall (PR) with and without the real-time navigation guidance during the scan to demonstrate the possible benefits of our tool.
All datasets used, are proprietary of the group.

We conducted all of the tests using a calibrated Olympus CF-H185L/I colonoscope on a 3D printed colon model. The colon model was manufactured by segmenting a CT colon scan from~\cite{colonographydata} and post-processing it to recover the 3D structure of the colon.  The model was fabricated from the final mesh using a 3D printer. ColNav was run on a high-performance computer equipped with an AMD Ryzen 3960x processor, 128 GB of RAM, and an NVIDIA A6000 GPU. The algorithm ran at a speed of 20 FPS while the live endoscope stream was in its native frequency, enabling real-time usage and guidance during the scans.



 


The annotations for the first test, and the scans in the second test were performed by physicians who, on average, had 6.5 years of experience and conducted 5000 colonoscopies. We also used the baseline PR experiment (without ColNav) as a standard, and only included physicians with a recall of over $50\%$.

\textbf{Colon unfolding verification:} 
Two physicians were asked to annotate the coverage level of 83 short clips captured using our colon model. To assess the annotators agreement, we used weighted Cohen's kappa coefficient~\cite{kappa}, due to the ordinal nature of the coverage categories. 
The resulting weighted kappa score of approximately $60\%$ indicated a "moderate" level of agreement.

However, the absolute coverage value given by a single physician might be subjective, making calibration and comparison difficult. Thus, to overcome this issue, we tested for agreement on the relative score of the four quadrants. We used Cohen's kappa to measure the agreement between the physicians on the order of the sorted quadrants based on their coverage score (most covered quadrant is first, least one is last). Cohen's kappa using the relative coverage scores between the two annotators is $84.7\%$ meaning 'almost perfect agreement', which showcases that this comparison method is more suited for the task.

Based on this approach, we applied ColNav to compute a flattened image for each short clip. Each flattened image was partitioned into four quadrants, and the coverage percentage was calculated for each quadrant. To evaluate our predictions, we mapped the coverage percentages to three categories using a simple threshold: ($coverage <= 60\%$: 'mostly not covered' , $60\% < coverage <= 80\%$: 'partially covered', $80\% < coverage <= 100\%$: 'mostly covered'). The results shown in Table~\ref{tab:kappa} present ColNav high levels of agreement with the two physicians, with agreement rates of $88.4\%$ and $85.9\%$ respectively. These results demonstrate that ColNav accurately represents the scanned colon and has high inter-rater reliability with physicians for predicting coverage levels.


\begin{table}[tbp]
\small
\begin{minipage}{0.52\textwidth}
\centering
 \caption{Weighted Cohen's Kappa over the relative coverage scores.}
\label{tab:kappa}
\begin{tabular}{|l|c|}
\hline
  & Cohen's Kappa[$\%$] \\ \hline
Annotators A, B &$84.7$ \\ \hline 
Anno. A,  \textbf{ColNav} &\textbf{88.4}\\ \hline 
Anno. B, \textbf{ColNav} &\textbf{85.9}\ \\ \hline
\end{tabular}
\end{minipage} 
\begin{minipage}{0.43\textwidth}
\centering
 \caption{Polyp Recall \& Coverage with/out ColNav ($avg. \pm std$)}
\label{tab:BDR}
\begin{tabular}{|l|c|c|}
\hline
  & PR[$\%$] & Cov.[$\%$] \\ \hline
Without ColNav & $77.8 \pm 3.9$ & $91.6 \pm 1.5$ \\ \hline 
With \textbf{ColNav} & \textbf{88.9} $\pm 3.9$ & \textbf{96.4} $\pm 1.0$\\ \hline 
\end{tabular}
\end{minipage}

\end{table}

\begin{figure}[hptb]
    \centering
    \begin{tabular}[t]{cc}
        \begin{subfigure}{.45\textwidth}
                \begin{tabular}{c}
                    \begin{subfigure}[t]{1\textwidth}
                        \centering
                        \includegraphics[width=1\textwidth]{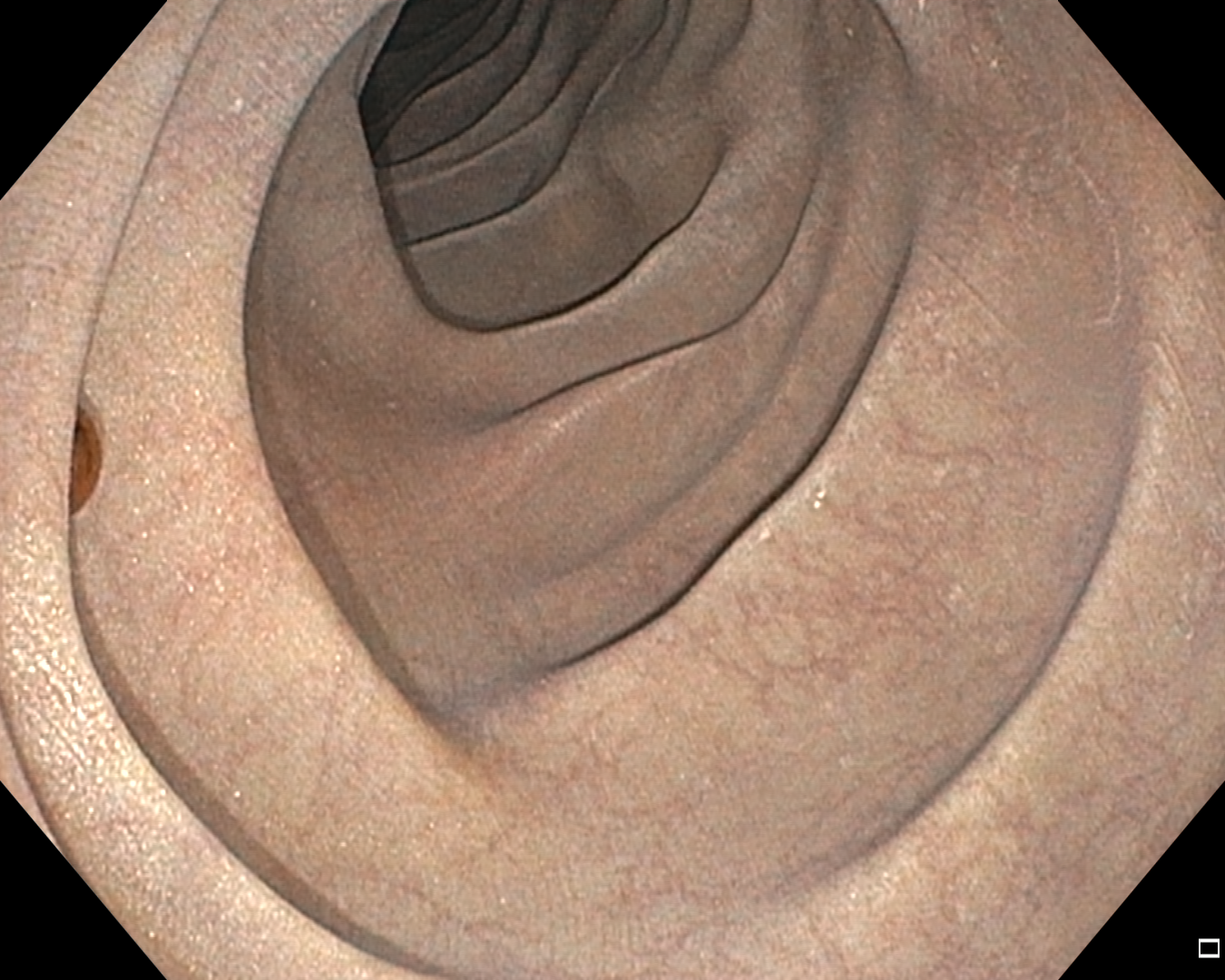}
                    \end{subfigure}\\
                    \begin{subfigure}[t]{1\textwidth}
                        \centering
                        \includegraphics[width=1\textwidth]{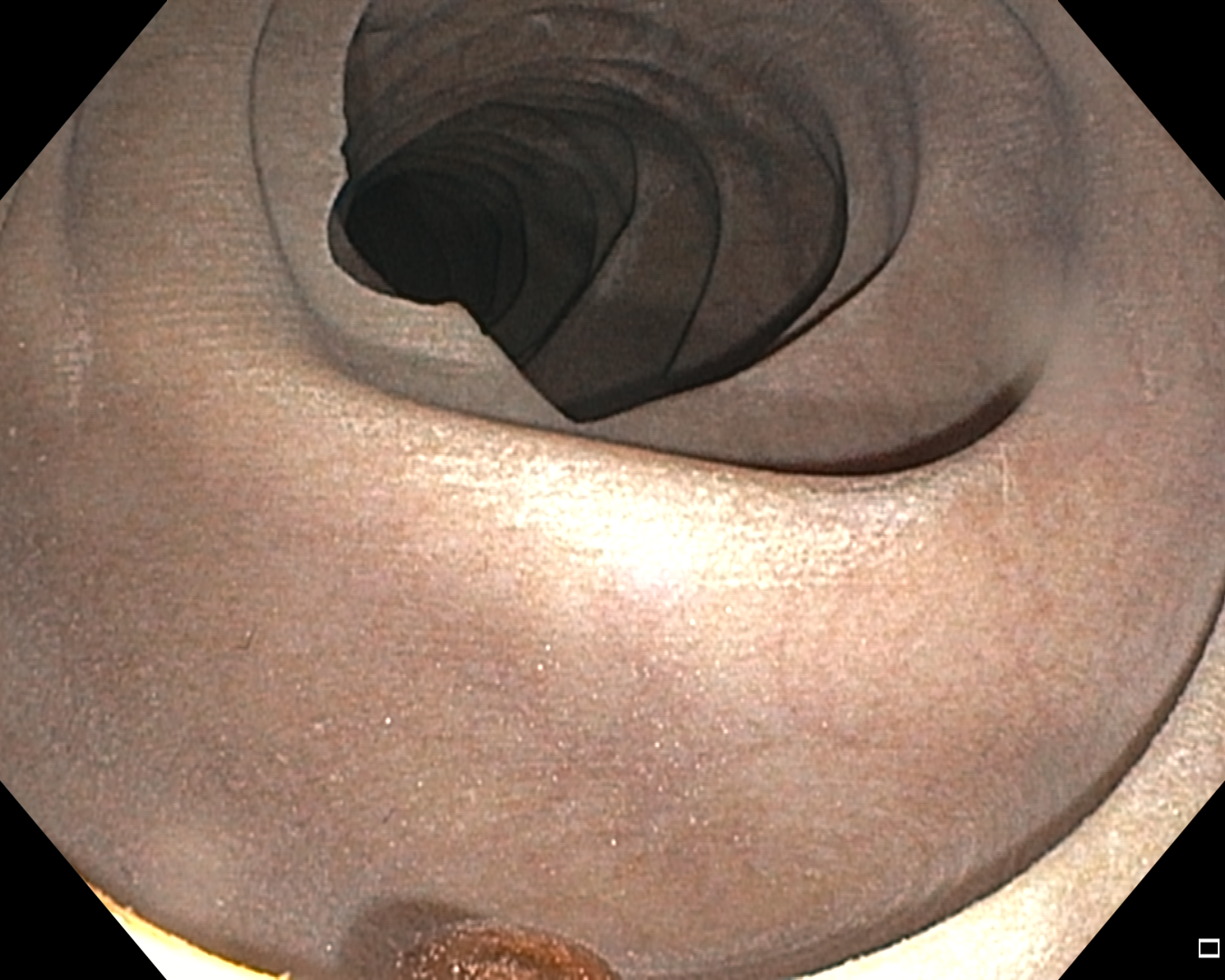}
                    \end{subfigure}\\
                    \begin{subfigure}[t]{1\textwidth}
                        \centering
                        \includegraphics[width=1\textwidth]{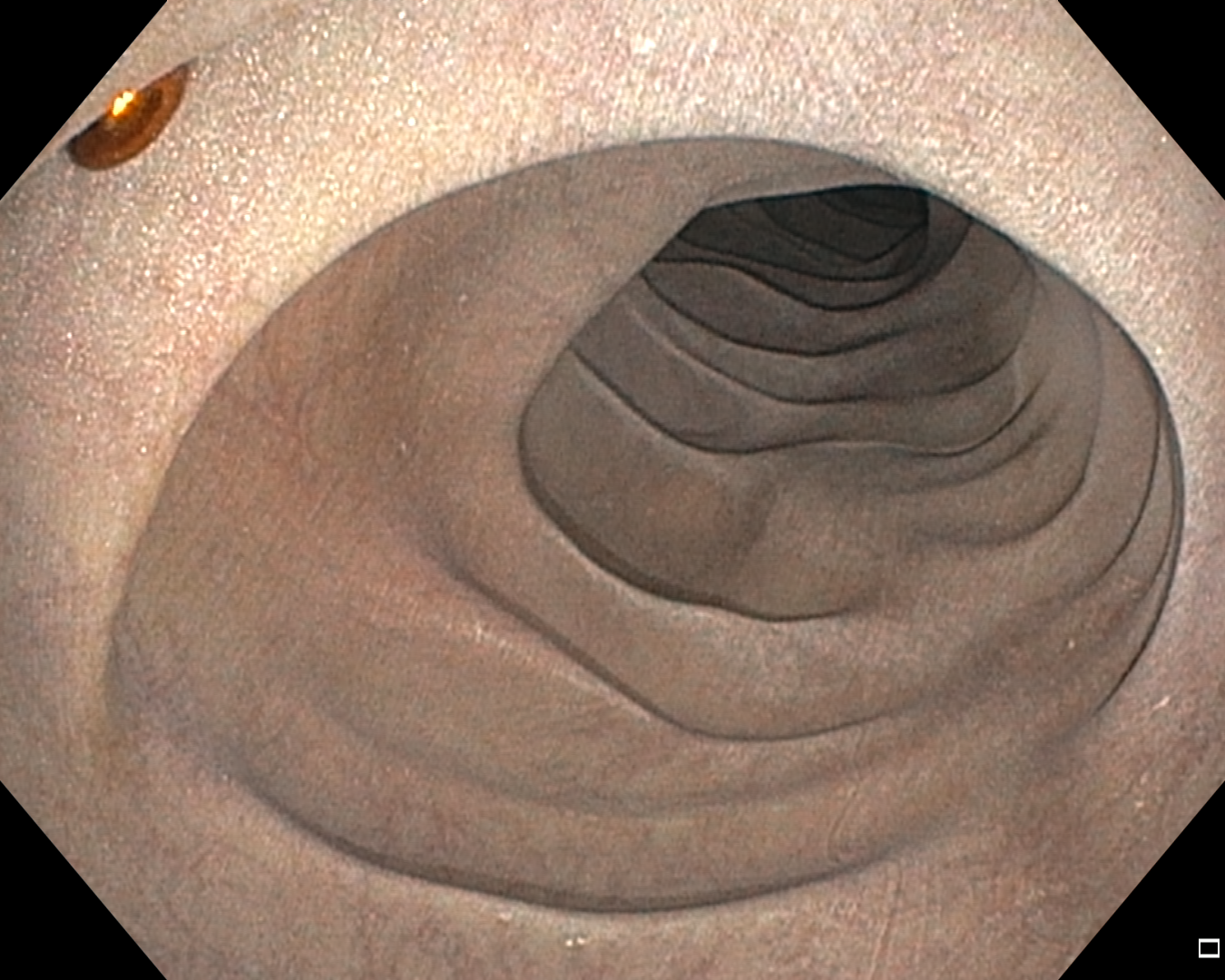}
                    \end{subfigure}\\

            \end{tabular}
        \end{subfigure} &
        \begin{subfigure}{.45\textwidth}
                \begin{tabular}{c}
                    \begin{subfigure}[t]{1\textwidth}
                        \centering
                        \includegraphics[width=1\textwidth]{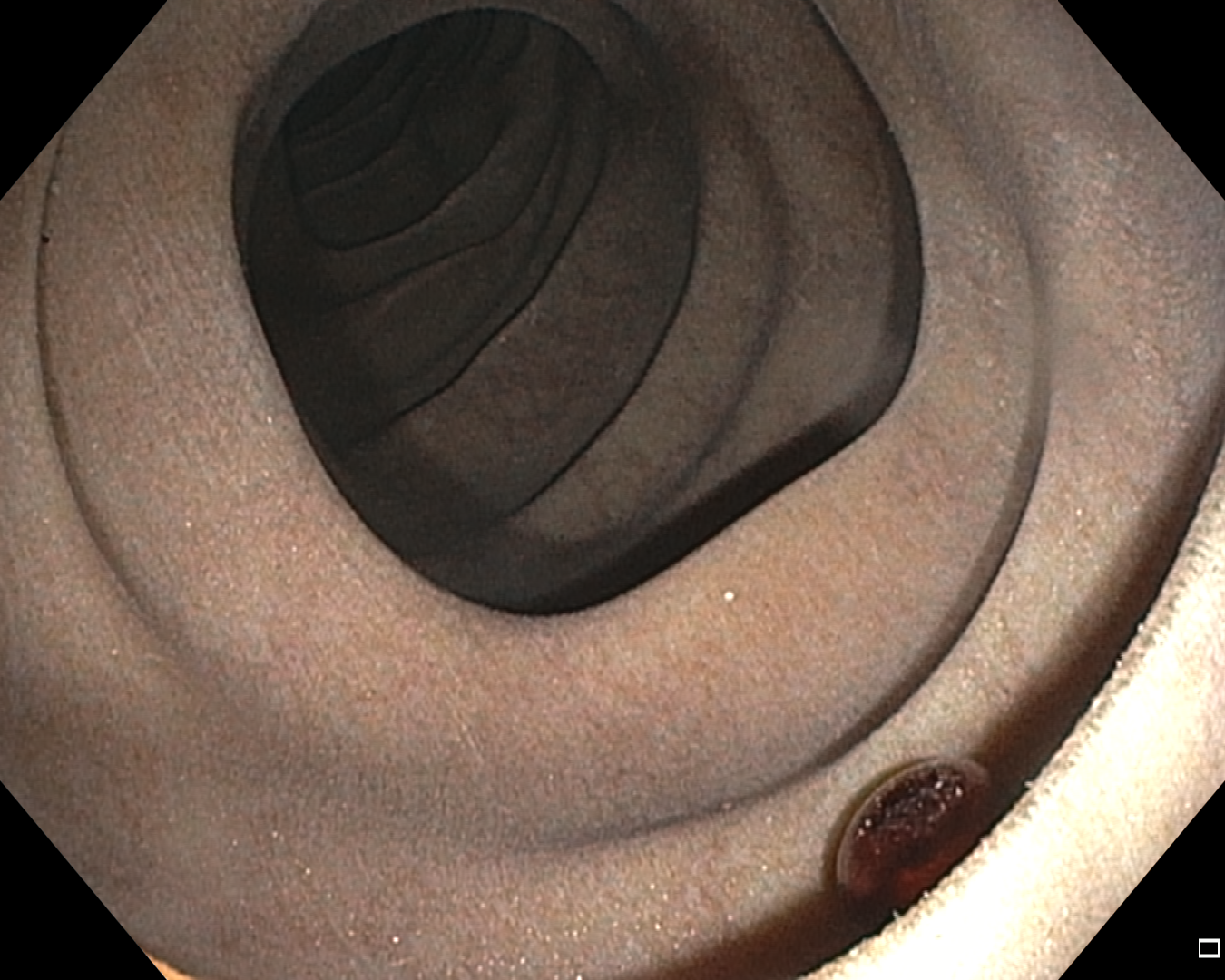}
                    \end{subfigure}\\
                    \begin{subfigure}[t]{1\textwidth}
                        \centering
                        \includegraphics[width=1\textwidth]{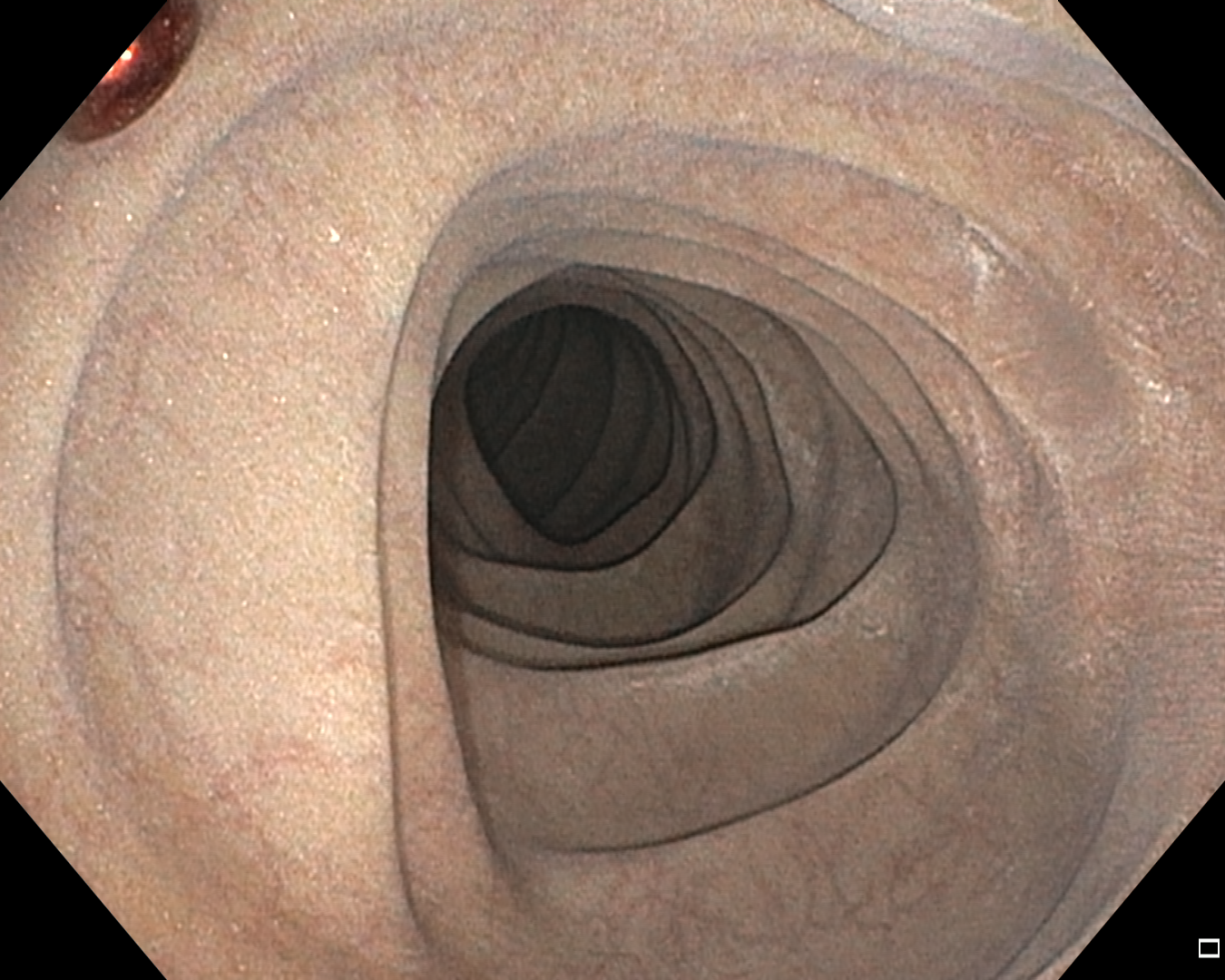}
                    \end{subfigure}\\
                    \begin{subfigure}[t]{1\textwidth}
                        \centering
                        \includegraphics[width=1\textwidth]{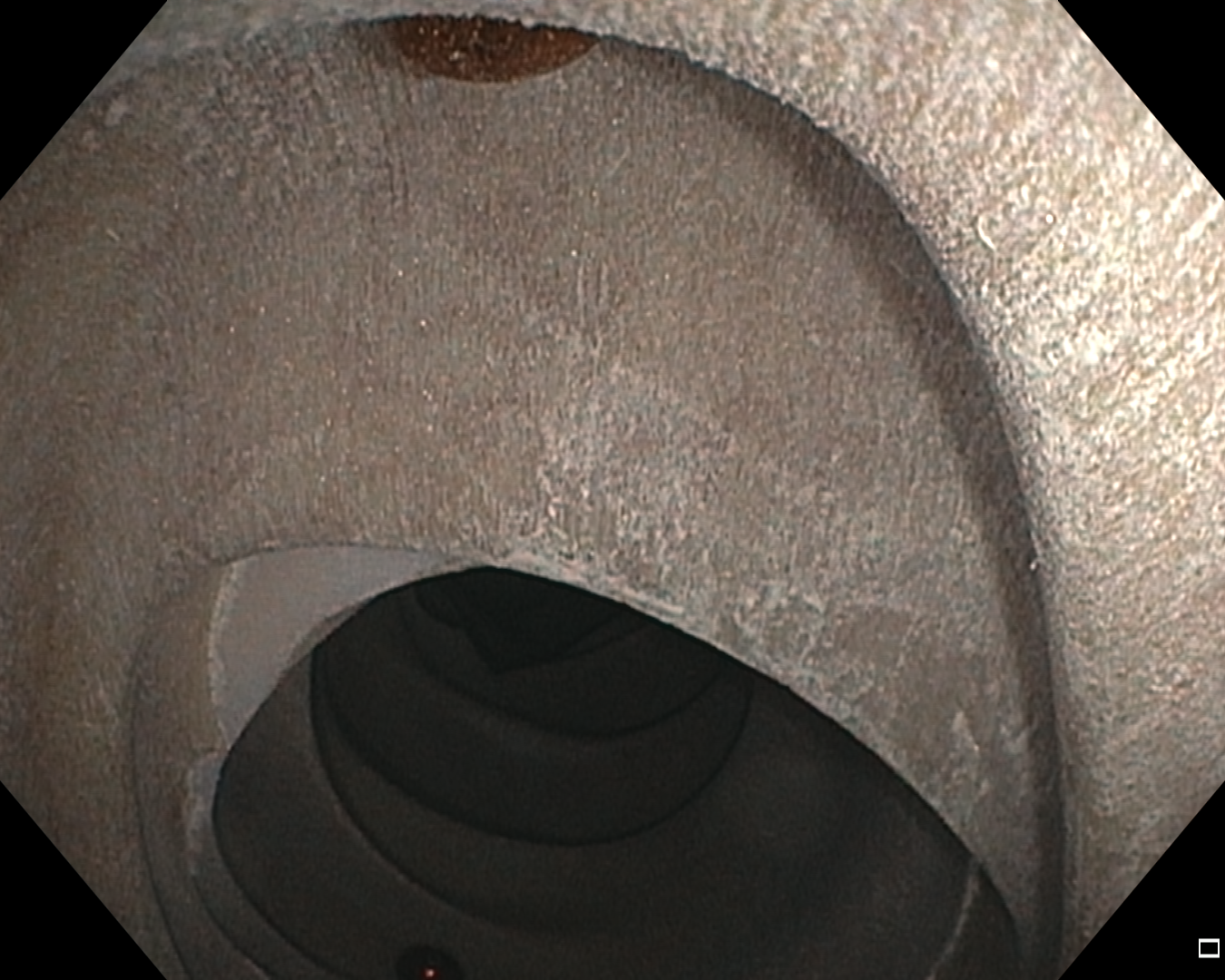}
                    \end{subfigure}\\

            \end{tabular}
        \end{subfigure} 
    \end{tabular}
\caption{frames containing hidden magnetic balls in multiple colors, that were used to simulate polyps.}
\label{Supp:ball}
\end{figure}

\textbf{Polyp Recall Impact:}
Real-time estimation of coverage offers the crucial advantage of guiding physicians to potentially missed areas and enhancing the detection of polyps in these regions. To evaluate this capability, we conducted a simulation study by concealing $18$ small magnetic balls (diameter=$5 mm$) within our colon model to simulate polyps (see Fig.~\ref{fig:exps}, Supplementary Fig~\ref{Supp:ball}). Three trained physicians were recruited to perform an  optical colonoscopy on the model, with and without ColNav, while recording their coverage and recall, i.e. the ratio between the number of balls detected during each test and the total number of balls. Scans were conducted in the same manner, starting from the end of the model ('cecum'), the colon was examined while the endoscope was withdrawn. To prevent location bias, we used balls of multiple colors and assigned each physician a different combination of colors in each test phase (with/without ColNav). The results, as presented in Table~\ref{tab:BDR}, reveal that physicians using ColNav achieved $11.1\%$ higher polyp recall (PR) and $4.8\%$ better coverage, demonstrating the effectiveness of our solution and supporting our belief that ColNav could improve PDR in clinical scenarios.

\textbf {Real colonoscopy videos:} ColNav was used in an offline manner on short clips of recorded procedures. See Supplementary Fig.~\ref{Supp:real} for example of the output.

\section{Conclusion}
We have presented ColNav, the first of its kind real-time colon navigation system, which not only calculates coverage, but also provides augmented guidance to un-surveyed areas without disrupting the procedure. The coverage estimation has been shown to have high correlation with experts. Using the system, physicians were able to improve their coverage and recall in detecting findings within the colon. The system was qualitatively evaluated offline on real-life procedures. Further research will focus on improving real-time performance and robustness to extreme colon deformation using non-rigid SLAM. 
\bibliographystyle{splncs04}

\bibliography{bib/DaiNZIT16.bib}

\clearpage
\pagebreak 


\section{Supplementary}

\begin{figure}[h!tb]
    \centering
    \begin{tabular}[t]{cc}
        
        \begin{subfigure}{.44\textwidth}
                \begin{tabular}{c}
                \smallskip
                    \begin{subfigure}[t]{1\textwidth}
                        \centering
                        \includegraphics[width=0.81\textwidth]{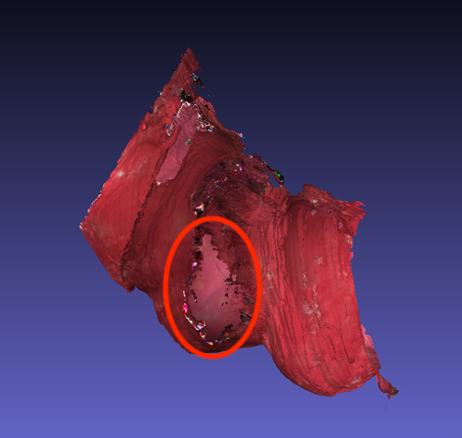}
                    \end{subfigure}\\
                    \begin{subfigure}[t]{1\textwidth}
                        \centering
                        \includegraphics[width=0.81\textwidth]{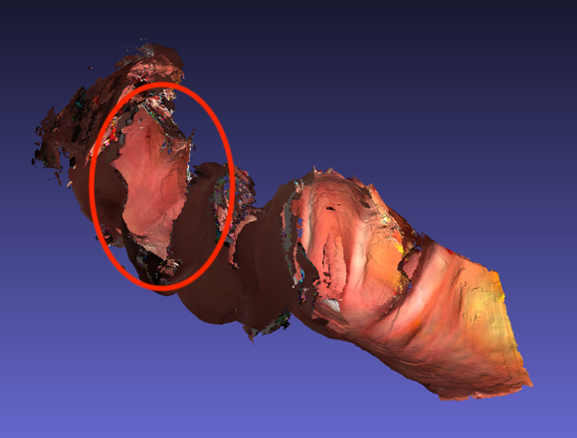}
                    \end{subfigure}\\
                                        \begin{subfigure}[t]{1\textwidth}
                        \centering
                        \includegraphics[width=0.81\textwidth]{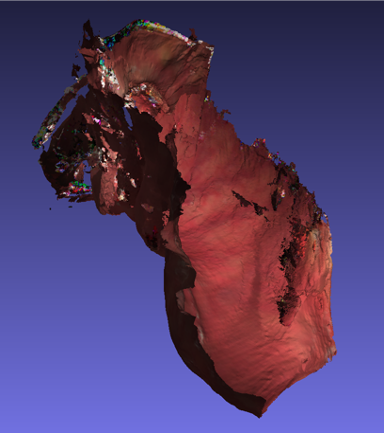}
                    \end{subfigure}

            \end{tabular}
            \caption{reconstructed mesh}

        \end{subfigure}
        
        \begin{subfigure}{.24\textwidth}
                \begin{tabular}{c}
                \smallskip
                    \begin{subfigure}[t]{1\textwidth}
                        \centering
                        \includegraphics[height=1.413\textwidth, width=0.75\textwidth]{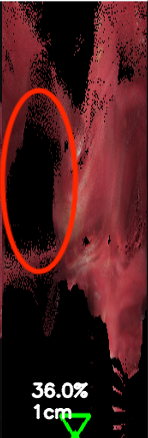}
                    \end{subfigure}\\
                    \begin{subfigure}[t]{1\textwidth}
                        \centering
                        \includegraphics[height=1.134\textwidth, width=0.75\textwidth]{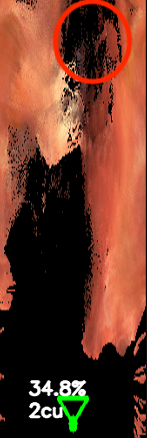}
                    \end{subfigure}\\
                                        \begin{subfigure}[t]{1\textwidth}
                        \centering
                        \includegraphics[height=1.665\textwidth, width=0.75\textwidth]{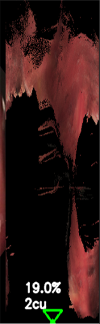}
                    \end{subfigure}

            \end{tabular}
            \caption{flattened image}
        \end{subfigure}

    \end{tabular}
    \caption{ColNav on real colonoscopy video segments. (a) 3D reconstructed mesh, (b) flattened image. Red ellipses mark corresponding uncovered areas. Best viewed in color.}
\label{Supp:real}
\end{figure}


\end{document}